\pdfoutput=1

\documentclass[11pt]{article}

\usepackage{acl}

\usepackage{times}
\usepackage{latexsym}
\usepackage{pifont}
\newcommand{\xmark}{\ding{55}}%

\usepackage[T1]{fontenc}

\usepackage[utf8]{inputenc}

\usepackage{microtype}
\usepackage{times}
\usepackage{latexsym}
\usepackage{xspace}
\usepackage{booktabs}
\usepackage{graphicx}
\usepackage{multirow}
\usepackage{tikz}
\usetikzlibrary{plotmarks}
\usepackage{fontawesome5}
\usepackage{xutfsym}
\usepackage[tight,footnotesize]{subfigure}

\usepackage{amsmath,amssymb,amsfonts,amsthm}
\usepackage{mathtools}
\usepackage{algorithm}
\usepackage{algorithmic}
\usepackage{soul}
\usepackage{enumitem}
\usepackage{todonotes}
\usepackage{xcolor}
\usepackage{wrapfig,lipsum,booktabs}
\usepackage{fontawesome5}
\usepackage{tikz}
\usetikzlibrary{plotmarks}
\theoremstyle{definition}
\newtheorem{definition}{Definition}[section]

\newcommand\marksymbol[2]{\tikz[#2,scale=1.2]\pgfuseplotmark{#1};}

\newcommand{\sysname}{\textsc{NANO}\xspace}

\newcommand{\draftonly}[1]{#1}

\newcommand{\draftcomment}[1]{\draftonly{#1}}
\newcommand{\jiao}[1]{\textcolor{magenta}{\tiny [#1]$_{\textrm{jiao}}$}}
\newcommand{\sg}[1]{\draftcomment{\textcolor{orange}{\small [#1]$_{\textrm{sg}}$}}}
\newcommand{\ttvu}[1]{\textcolor{blue}{\bf \small [#1 --Tu]}}

\newcommand{\jacob}[1]{\textcolor{blue}{\bf \small [#1 --Jacob]}}
\newcommand{\outline}[1]{\textcolor{purple}{\bf \small [Outline: #1]}}
\newcommand{\comet}{\textsc{COMET}\xspace}
\newcommand{\prism}{\textsc{Prism}\xspace}
\newcommand{\yisi}{\textsc{YiSi}\xspace}
\newcommand{\bleu}{\textsc{BLEU}\xspace}
\newcommand{\chrf}{\textsc{chrF}\xspace}
\newcommand{\bleurt}{\textsc{BLEURT}\xspace}

\makeatletter
\newcommand*{\rom}[1]{\expandafter\@slowromancap\romannumeral #1@}
\makeatother
\newcommand{\say}[1]{``\textit{#1}''}

\newcommand{\mkcleancomments}{
	\renewcommand{\jiao}[1]{}
	\renewcommand{\sg}[1]{}
	\renewcommand{\ttvu}[1]{}
	\renewcommand{\jacob}[1]{}
	\renewcommand{\outline}[1]{}
}
\mkcleancomments

\usepackage[most]{tcolorbox}

\definecolor{darkyellow}{rgb}{1.0, 0.85, 0.4}
\definecolor{lightblue}{rgb}{0.70, 0.89, 1.0}
\definecolor{lightgrey}{rgb}{0.60, 0.60, 0.60}
\definecolor{lightpurple}{rgb}{0.91, 0.71, 0.99}
\definecolor{myblue}{rgb}{0.0, 0.44, 1.0}
\definecolor{mygreen}{rgb}{0.55, 0.71, 0.0}

\newtcbox{\inlineyellowbox}[1][]{enhanced, box align=base, nobeforeafter, colback=darkyellow, colframe=white, size=small, left=0pt, right=0pt, boxsep=0pt, #1}
\newtcbox{\inlinebluebox}[1][]{enhanced, box align=base, nobeforeafter, colback=lightblue, colframe=white, size=small, left=0pt, right=0pt, boxsep=0pt, #1}
\newtcbox{\inlinepurplebox}[1][]{enhanced, box align=base, nobeforeafter, colback=lightpurple, colframe=white, size=small, left=0pt, right=0pt, boxsep=0pt, #1}

\definecolor{WowColor}{rgb}{.75,0,.75}
\definecolor{SubtleColor}{rgb}{0,0,.50}

\newcounter{margincounter}

\newcommand{\yd}[1][d]{\ensuremath y^{(#1)}}

\title{Dialect-robust Evaluation of Generated Text}

\author{
Jiao Sun$^{1,2}$\thanks{\xspace\xspace Jiao and Tu are interns at Google.} \, \textbf{Thibault Sellam}$^1$ \, \textbf{Elizabeth Clark}$^1$ \, \textbf{Tu Vu}$^{1,3*}$ \, \textbf{Timothy Dozat}$^1$ \, 
\\ 
\textbf{Dan Garrette}$^1$ \, \textbf{Aditya Siddhant}$^1$ \, \textbf{Jacob Eisenstein$^1$ \, Sebastian Gehrmann}$^1$ 
\\
$^1$Google Research 
\\ 
$^2$University of Southern California \, $^3$University of Massachusetts Amherst 
\\
\texttt{\{jiaosun,tsellam,eaclark,ttvu,tdozat,dhgarrette,adisid\}} 
\\
\texttt{\{jeisenstein,gehrmann\}@google.com}
}

\begin{document}

\maketitle

\begin{abstract}
Evaluation metrics that are not robust to dialect variation make it impossible to tell how well systems perform for many groups of users, and can even penalize systems for producing text in 
lower-resource dialects.
However, currently, there exists no way to quantify how metrics respond to change in the dialect of a generated utterance. 
We thus formalize \emph{dialect robustness} and \emph{dialect awareness} as goals for NLG evaluation metrics.
We introduce a suite of methods and corresponding statistical tests one can use to assess metrics in light of the two goals. 
Applying the suite to current state-of-the-art metrics, we demonstrate that they are not dialect-robust and that semantic perturbations frequently lead to smaller decreases in a metric than the introduction of dialect features.
As a first step to overcome this limitation, we propose a training schema, \sysname, which introduces regional and language information to the pretraining process of a metric. 
We demonstrate that \sysname provides a size-efficient way for models to improve the dialect robustness while simultaneously improving their performance on the standard metric benchmark.
\end{abstract}

\section{Introduction}
Most natural language generation (NLG) evaluation metrics compare a system output against a human-written reference. References are usually drawn from a relatively narrow range of linguistic styles. They often exclude varieties like Indian English or Iberian Portuguese, which are \emph{geographical dialects}
with millions of speakers. As a result, outputs in dialects that are not represented in the reference may score poorly, discouraging the development of systems to meet the needs of these language communities. Although contemporary metrics such as \comet \citep{comet} can be reference-free, they still rely on training data and rater pools that do not cover all dialects of interest, leading to a high number of out-of-domain dialects. The performance of evaluation metrics on these out-of-domain dialects has not been quantified.

We define a \emph{dialect-robust} evaluation metric as one that produces the same score for system outputs that share the same semantics, but are expressed in different dialects. To understand whether current evaluation metrics are dialect-robust, we propose to quantify the dialect robustness at the dialect feature-level and sentence-level. The analyses  measure the dialect-sensitivity of evaluation metrics by comparing semantics-preserving dialect edits to perturbations that change the meaning of sentences.

Through our analyses, we demonstrate that multiple state-of-the-art NLG evaluation metrics are not robust to dialects of Mandarin, English, and Portuguese. In many cases, system outputs that are perturbed so as to differ semantically from the reference score higher than outputs in which the only change is to the dialect.
With the goal of increasing the dialect robustness and without performance degradation on standard benchmarks, we propose a training schema \sysname. \sysname is an unsupervised pretraining step to a metric that distills dialect information of the multilingual pretraining dataset into a model, which we demonstrate leads to improved dialect robustness.

Based on our findings, we lay out research goals toward dialect-inclusive metrics. Moving beyond dialect robustness, we formalize the goal of \textit{dialect awareness}, in which metrics can be applied to any user-specified language and dialect regardless of the language of the reference or source document.

\begin{figure*}
    \centering
    \includegraphics[width=\linewidth]{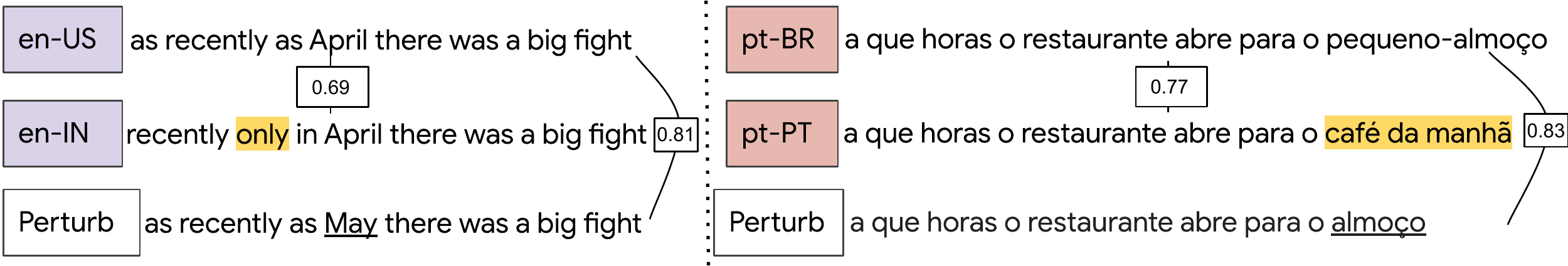}
    \caption{An illustration of dialect robustness in the context of generation evaluation. We define \emph{dialect robustness} as evaluation metrics that are expected to have the same output across dialects that share the same semantics. Dialect edits (highlighted in \inlineyellowbox{yellow}) should not lead to a greater degradation of score than edits that change the underlying semantics (highlighted in \underline{underline}). BLEURT-20 in the figure assigns higher score to semantically-perturbed sentences than sentences with dialect features, exposing its vulnerability to dialects.}
    \label{fig:illustration}
\end{figure*}

\section{Dialect Robustness and Awareness}
\label{sec:bg}
Dialects can be regarded as linguistic subdivisions that align with communities of speakers, often grouped by geographical or demographic attributes~\citep{chambers1998dialectology}. 
A classic example is nation-level varieties, such as Brazilian and Iberian Portuguese.
Dialects are distinguished from each other by a set of \emph{dialect features}, which can operate at the levels of  pronunciation, lexicon, rhetorical devices, and grammar~\citep{whiteman2013writing}; one working definition of dialect is as a set of correlated features~\citep{Nerbonne2009DataDrivenD}. 

Two examples of dialect features are shown in \autoref{fig:illustration}. The left side shows the English dialect feature ``focus \emph{only}'', which distinguishes Indian English from other varieties, such as US English~\citep{lange2012syntax}. The feature changes the surface form but not the underlying semantics. 
The right panel of \autoref{fig:illustration} shows the Portuguese dialect feature of different lexical choice for the same semantics (``breakfast''), which distinguishes Iberian Portuguese from Brazilian Portuguese. Many dialect features are acceptable in multiple dialects: for example, zero definite article (``$\varnothing$ main reason is \ldots'')\footnote{\url{https://ewave-atlas.org/parameters/62\#2/7.0/7.9}} is used in Indian English, Singapore English, and several other post-colonial dialects.

\paragraph{Dialect Robustness} Consider a translation system that produces Iberian Portuguese outputs at a task where it is desirable to generate text in a variety of dialects. If all the training data for the metric used to evaluate generation quality comes from Brazilian Portuguese, it will likely assign a lower score to Iberian Portuguese outputs, thereby misrepresenting system quality and disincentivizing further development of the more diverse system in favor of one that only produces Brazilian Portuguese. 
To be able to measure this effect, we define dialect robustness in the context of NLG evaluation as:
\begin{definition}[Dialect robustness]
Let $\yd$ and $\yd[d']$ be two system outputs that are semantically equivalent but written in different dialects. An evaluation metric $m: \mathcal{Y} \to \mathbb{R}$ is \textbf{dialect robust} iff $m(\yd) = m(\yd[d'])$ for all such $(\yd, \yd[d'])$.\footnote{For simplicity we do not include the reference in this definition. A corpus-level reference-based metric could be defined as $\frac{1}{N} \sum_i m_i(y_i)$ with $m_i(y_i) = \delta(y_i, r_i)$, with $r_i$ indicating the reference for example $i$ and $\delta : \mathcal{Y} \times \mathcal{Y} \to \mathbb{R}.$ Similarly, a corpus-level quality estimation metric could be defined with $m_i(y_i) = \delta(y_i, x_i)$ with $x_i$ indicating the input, such as the source language or passage to be summarized. For the corpus-level metric to be dialect robust (or $\phi$-robust), all $m_i$ must be dialect robust (or $\phi$-robust).}
\end{definition}
This definition is strict: it would not apply to any system that produced even small differences in score between semantically equivalent, regionally distinct outputs. For that reason, we propose a relaxed criterion, which compares the change in the metric induced by dialect to changes induced by semantic perturbations:
\begin{definition}[$\phi$-Dialect robustness]
\label{def:phi-robustness}
Let $\yd$ and $\yd[d']$ be two semantically-equivalent system outputs that differ in dialect. Let $\phi: \mathcal{Y} \to \mathcal{Y}^*$ be a semantic perturbation function that maps an input to a set of outputs whose semantics are different from the input. An evaluation metric $m: \mathcal{Y} \to \mathbb{R}$ is $\phi$-dialect robust if $m(\yd, \yd[d']) > m(\yd, \tilde{y})$ for all semantically-equivalent $(\yd, \yd[d'])$ and all $\tilde{y} \in \phi(\yd)$.\end{definition}

\paragraph{Dialect Awareness} Consider a translation system that is supposed to translate into Brazilian Portuguese but instead produces Iberian Portuguese. In this case, a dialect-robust metric is undesirable because it is unable to detect this mistake. To account for these cases, we define dialect awareness:
\begin{definition}[Dialect-awareness]
\label{def:dialect-awareness}
Let $\mathcal{T}$ be a set of dialect tags. A metric ${m: \mathcal{Y} \times \mathcal{T} \to \mathbb{R}}$ is \textbf{dialect aware} iff $m(\yd, d) \geq m(\yd[d'], d)$ for all semantically-equivalent input pairs $(\yd, \yd[d'])$ where $\yd$ is in dialect $d \in \mathcal{T}$ and $\yd[d']$ is in dialect $d' \neq d$.
\end{definition}

\noindent Informally, a metric is dialect aware if, given a dialect identifier and a pair of semantically-equivalent texts that vary by dialect, it assigns the highest score to the text in the dialect specified by the identifier. Dialect awareness is undefined with respect to inputs that are not semantically equivalent. This means that the definition is agnostic as to whether the metric should prioritize matching the target semantics or the target dialect.

\autoref{fig:illustration} illustrates the concepts of dialect robustness and dialect awareness. The top two rows of each panel vary only by dialect; the bottom row shows semantic perturbations of the top row. $\phi$-dialect robustness implies that the top row is scored as more similar to the middle row than to the bottom row. Dialect awareness implies that the quality of the surface form in each row should be highest when paired with the correct dialect label.

\paragraph{Is Semantic Equivalence Realistic?} 
The above definitions presume that it is possible to characterize utterances in different dialects as semantically equivalent. Such characterizations have been criticized as lacking a strong foundation for semantic equivalence, outside the limited case in which the dialect differences are purely phonological~\citep{lavandera1978does,romaine1981problem}. One such criticism is that a pair of utterances might be semantically equivalent for some communicative purposes, but not for others. To avoid the gray area between dialect differences that change semantics and those that do not, we design perturbations that have a small surface-level impact on the original utterance but a strong effect on its meaning, e.g.~by negating the main proposition or changing an important semantic argument. This establishes a necessary but not sufficient condition for dialect robustness: if a metric scores such perturbations more highly than dialect pairs, then it is certainly not dialect robust. Proving that a metric is dialect robust is more difficult, because it requires constructing more subtle semantic perturbations that are harder to distinguish (even conceptually) from dialect variables. Furthermore, from a practical standpoint we cannot evaluate $\yd$ with respect to \emph{all} semantic perturbations $\tilde{y} \in \phi(\yd)$, but the existence of perturbations for which $m(\yd, \tilde{y}) > m(\yd, \yd[d'])$ is enough to disprove dialect robustness.

\section{Existing Metrics}

To assess the quality of a generated text, most automatic evaluation approaches compare it to a ``ground truth'' reference, with higher similarity to the reference implying higher-quality output~\citep{DBLP:journals/corr/abs-2006-14799}. Similarity can be based on lexical features or distributed representations. When distributed representations are used, they may be unsupervised~\citep{DBLP:conf/iclr/ZhangKWWA20} or fine-tuned on a corpus of human ratings.
In addition to these similarity-based metrics, there are also reference-free metrics for quality estimation (e.g., \comet-QE), which we discuss in \S\ref{sec:finetune}.

\paragraph{Lexical Evaluation Metrics} Many evaluation metrics including \bleu~\citep{bleu} and \chrf~\citep{chrf} use lexical features such as n-gram overlap to measure similarity and remain popular evaluation metrics because of their lightweight and fast computation.
However, due to their reliance on surface forms, \bleu and \chrf have limited robustness to superficial syntactic differences between system outputs and references~\citep{bleurt}. 
As dialects inherently include lexical variables, traditional evaluation metrics based on lexical overlap are expected to not perform well in terms of dialect robustness. 

\paragraph{Distributed Evaluation Metrics}
Moving beyond surface forms, recent advances such as \bleurt~\citep{pu-etal-2021-learning},\footnote{We use BLEURT-20 checkpoint from \citet{pu-etal-2021-learning} different from the original BLEURT~\cite{bleurt}.} and \comet leverage the representations from models that are trained on human ratings. \bleurt pretrains RemBERT~\cite{Chung2021RethinkingEC} on augmented data from Wikipedia and then finetunes on human ratings from WMT corpora. \comet is trained on the mixture of WMT and another two corpora, QT21~\cite{QT21} and MQM~\cite{freitag-etal-2021-experts} which both rely on machine translated outputs. \prism is trained on generated paraphrases from  a mixture of data resources in 39 languages and does not require human ratings during training. \yisi directly utilizes the multilingual representation from multilingual BERT~\citep{devlin-etal-2019-bert} for scoring. In summary, existing learned metrics either utilize the multilingual representation from pretrained models, or create multilingual training data through various augmentation strategies. However, none of them explicitly accounts for dialectal variations during training. 

\section{Testing Dialect Robustness}
\label{sec:method}

\outline{Test NLG evaluation metrics' ability on both dialect robustness and the performance on standard WMT metrics benchmark.}

In this section, we describe our methodology for assessing dialect robustness. We first introduce two ways to perturb sentences to get two comparable metrics' outputs and then describe the statistical tests we use to aggregate the outputs over a corpus.

\subsection{Micro-level Dialect Features}
\label{sec:micro}

\newcommand{\en}[0]{\ensuremath \textsc{en}}
\newcommand{\score}[0]{\ensuremath \sigma}
Dialect features are local edits that distinguish dialects while avoiding changes to the meaning of the text; an orthographic example is the spelling of the suffix \say{-or} vs \say{-our}, which distinguishes U.S.~vs U.K.~English. Our first robustness assessment uses such features. We start with a base sentence $y^{(\text{base})}_i$ taken from a corpus of sentences $D= \{y_1,\ldots,y_n\}$. We further assume access to a version of the same sentence in which a dialect feature was introduced, denoted $y^{(\text{dialect})}_i$. Following Definition \autoref{def:phi-robustness}, we introduce a semantic perturbation that changes $\yd[\text{base}]_i$ to  $\yd[\text{perturb}]_i$. 
Again using English as an example, from the US English base sentence \say{as recently as April\ldots}, we may produce the Indian English  version \say{recently only in April\ldots} (using the feature \emph{focus-only}), and the semantic perturbation \say{as recently as May\ldots}. 
 
Let $m(y_i, y_j)$ be a metric function that takes a candidate sentence $y_i$ and a reference $y_j$ as input, and produces a score $\score$. Given the above defined variations of $y_i$, we define the dialect and perturbed scores as

\begin{align}
\score^{(\text{dialect})}_{m, i} &=   m(y^{(\text{dialect})}_i, y^{(\text{base})}_i) \\
\score^{(\text{perturb})}_{m, i} &=  m(y^{(\text{perturb})}_i, y^{(\text{base})}_i).
\end{align}
To satisfy Definition \ref{def:phi-robustness}, $\score^{(\text{dialect})}_{m, i}$ should score higher than $\score^{(\text{perturbation})}_{m, i}$ across the sentences in the corpus. This implies, as a necessary but not sufficient condition, that $\mathbb{E}_{i\sim D}[\score^{(\text{dialect})}_{m, i}] > \mathbb{E}_{i \sim D}[\score^{(\text{perturb})}_{m, i}]$. 

We consider three perturbation types: deletion, replacement and insertion. Each perturbation aims to change the sentence by only a single word or phrase, so as to induce a strong semantic change with a minimal impact to the surface form. Such perturbations are expected to yield challenging but clear comparisons against dialect variation.

There are no standard techniques for introducing semantic perturbations, so we apply fewshot-learning by prompting LaMDA~\citep{lamda}. 
For each perturbation type, we provide five exemplars and then prompt LaMDA for automatic semantic perturbation given a sentence $\yd[\text{en-base}]_i$.\footnote{Exemplars are shown in \autoref{app:five_shots}.}
Some sentences are not amenable to all perturbations --- for example, some are too short to support deletion --- so we choose one perturbation per sentence, with the preference order of replacement, insertion and then deletion, determined by the success rate of having a different sentence as output. 

\subsection{Sentence-level Dialect Rewrites}
\label{sec:sentence-level}

Micro-level dialect features require significant linguistic expertise to identify and have been defined for only a few languages. We thus introduce a less granular method that is based on parallel human translations. 
Given an English base sentence  $\en_i$, we obtain human translations $\yd[j]_i$ and $\yd[k]_i$ in dialects $j$ and $k$ of the target language, e.g., Brazilian and Iberian Portuguese. We can again use the metric $m$ to score the pair
\begin{align}
\score^{(\text{dialect})}_{m, i} &= m(y^{(j)}_i, y^{(k)}_i).
\end{align}

\noindent Because we have access to the English base sentence, we can use machine translation to generate a sentence in the target language $\en_i \xRightarrow[\text{MT}]{} \hat{y}_{i}^{(j^*)}$ which we can compare to, such that
\begin{align}
    \score^{(\text{MT})}_{m, i} &= m(y^{(j)}_i, \hat{y}_{i}^{(j^*)}).
\end{align}
\noindent Here, $j^*$ indicates the locale that we believe is most strongly targeted by the machine translation system (``pt-BR'' for Portuguese, ``zh-CN'' for Mandarin).

Finally, we construct target language perturbations by first perturbing the English source and then automatically translating:\footnote{While it is possible directly perturb the sentences in the target language, using the same English validated few-shot setup scales to more languages at the cost of a more English-centric perturbation style.}
\begin{align}
&    \en_i \xRightarrow[\text{perturbation}]{}  \tilde{\en}_i \xRightarrow[\text{MT}]{} \tilde{y}^{(j^*)}_i\\
&        \score^{(\text{perturb})}_{m, i} = m(y^{(j)}_i, \tilde{y}_{i}^{(j^*)}).
\end{align}

\noindent The perturbations are produced by prompting LaMDA with the same exemplars as in \S\ref{sec:micro}.

We expect $\mathbb{E}[\score^{(\text{MT})}_{m}] > \mathbb{E}[\score^{(\text{perturb})}_{m}]$, because both involve machine translation, but the latter also involves perturbation to the source. If we have $\mathbb{E}[\score^{(\text{perturb})}_{m}] >  \mathbb{E}[\score^{(\text{dialect})}_{m}]$ then metric $m$ strongly disprefers dialect variants, even in favor of inputs that are different in meaning due to the perturbation of the English source.

\subsection{Statistical Methods}
\label{sec:statistical}
As a necessary condition for dialect robustness, we test whether the expected scores for dialect rewrites exceed the expected scores for semantic perturbations. A challenge in correctly characterizing the uncertainty of these comparisons is that there is a substantial amount of variance over the original examples $\yd[\text{base}]_i$. We handle this with two styles of analysis:

\paragraph{Mixed-effect Regression} 

For metric $m$, example $i$, and condition $j \in \{\text{perturb, dialect, MT}\}$, we model the metric $\sigma_{m, i}^{(j)}$ via a mixed-effects regression~\citep{baayen2012mixed,speelman2018mixed},
\begin{align}
\sigma_{i}^{(j)} &= \theta_i + \phi_j + \epsilon_{i, j},
\label{eq:regression}
\end{align}
with the subscript $m$ implicit in each term. The first term $\theta_i$ is a random intercept associated with example $i$, which helps to address the variance across examples; $\phi_j$, the parameter of interest, is a fixed effect associated with the condition $j$; $\epsilon_{i, j}$ is a Gaussian error. Because all methods and conditions are applied to all examples, the predictors are uncorrelated. This makes it possible to interpret $\phi_{m, j}$ as an estimate of the expected change in the metric value corresponding to the application of metric $m$ in condition $j$. 

By including the $\theta_i$ term, the regression is conceptually equivalent to a pairwise comparison, in the sense that the regression also benefits from the additional power obtained by controlling for per-example variation.

\paragraph{Win/loss Analysis and Binomial Test} For a coarse-grained evaluation that is more easily comparable across metrics, we count how often each condition $j$ receives a higher score than condition $k$ in a pairwise comparison.
When $j$ represents dialect rewrites and $k$ represents semantic perturbations, a high win rate indicates that the metric is more likely to be dialect robust. To measure statistical significance, we apply a one-tailed binomial test, which computes the likelihood of achieving at least $n$ wins on $T$ trials given a null hypothesis win probability  $\frac{1}{2}$. In words, we test against the null hypothesis that for each example, a dialect rewrite and a semantic perturbation are equally likely to get the higher score.

As discussed in the next section, we perform multiple comparisons per metric, across different conditions and different languages. To adjust the $p$-values for multiple comparisons, we apply the Bonferroni correction~\citep{dror-etal-2017-replicability}.

\section{\sysname}
We hypothesize that explicitly encoding dialect information while pretraining a model will lead to an improved downstream robustness.  To test this hypothesis on learned metrics for text generation, we introduce \sysname,\footnote{The name is motivated by the dialect feature ``invariant tag (`isn't it', \underline{`no'}, \underline{`na'})'' ~\citep{lange2012syntax}.} a model-agnostic pretraining schema with the goal of improving dialect robustness without performance degradation on downstream metric benchmarks. 

\subsection{Acceptability Pretraining}
\label{sec:pretraining}

Given a pretrained model, we add a second pretraining phase to distill dialect information into the model.
Specifically, we define the \sysname-task as, given an expression $y_d$ in dialect $d$ which is part of a language $L$, identify whether $y_d$ is acceptable in a given dialect $d'$ or language $L'$.

\paragraph{Data}
To construct a training corpus for \sysname, we process mC4~\citep{mt5}. We split the corpus into sentences and use a Language Identification (LangID) model~\citep{Zhang2018AFC} by \citet{botha2017natural} to identify the language and locale information for the sentences.\footnote{We use a more current model that is capable of identifying the locale for Mandarin and Portuguese.}  Besides LangID output, mC4 provides the URL where a sentence originated from which we extract the region information as an indication of geographic dialect. For Portuguese and Mandarin, we filter an instance if the predicted locale does not agree with the region information from the URL. For other languages, we combine the LangID and region information as a noisy approximation for a dialect of the language in the specific region. For example, if the LangID model predicts that the language is English and the region in the URL indicates India (.in), we treat the instance as en-IN.\footnote{This is a noisy approximation because many dialects do not align with national borders. The development of a data-gathering approach for subnational and transnational dialects is an important topic for future work.} We compare three pretraining settings with an increasing noise: 1) Mandarin and Portuguese only; 2) Mandarin, Portuguese and selected English dialects and 3) ten languages with metric finetuning data evaluated during the WMT benchmark with ninety-five language variants following the classification by \citet{van2022writing}.\footnote{Appendix~\ref{app:lang_variants} provides the full list of WMT language variants, which does not cover Portuguese. Our reported results on PT shows \sysname's capability in a zero-shot setting.}  

Given a sentence, we balance the ratio of sampling a dialect or language tag using a parameter $\lambda$.  For instance, a sentence with gold dialect tag ``pt-BR'' can be a positive instance for the dialect itself or the general language ``pt-any''. 
At the same time, it can also be a negative instance for other dialect (e.g., ``en-IN'') or language (``en-any'').
The ratio of positive instances versus negatives instances is always 0.5 (for more discussion, see Appendix~\ref{app:ablation}.)

\paragraph{Modeling} We use mT5~\citep{mt5} as our base model because the model is pretrained on the mC4 dataset, matching with our corpus choice and ensuring tokenizer compatibility. During pretraining, we transform each sentence into the string \texttt{candidate}: \emph{\{sentence\}} \texttt{language}: \emph{\{language\_tag\}}, where the \emph{language\_tag} can be the dialect or the general language tag. The target label is zero or one, indicating whether the sentence belongs to the language tag. We adapt the Encoder-Decoder architecture of mT5 for regression by taking the logits of the first decoded token and applying the RMSE loss function between the logits and the label during model training. More details about model training is in Appendix~\ref{app:training}.

\subsection{Finetuning}
\label{sec:finetune}
Following the setup by \citet{pu-etal-2021-learning}, we use the test data from the WMT shared task from 2015 to 2019 as training data and use the WMT shared task 2020 as test data. Among previous works, \bleurt~\citep{pu-etal-2021-learning} and \yisi~\citep{yisi} are trained to measure the semantic similarity between candidate and reference within the same language. \comet, on the other hand, supports the cross-language quality estimation of a machine translation with or without reference, but does not support within-language assessment. To compare to all models, we finetune on all three settings following the input format in Appendix~\ref{app:input_format}.

\section{Experiments}
\label{sec:exp}
In this section, we demonstrate that existing metrics are not dialect robust by applying our proposed methods and statistical tests to existing corpora in English, Portuguese, and Mandarin (\S\ref{sec:dialect-robustness}). We show that language-aware pretraining via \sysname{} improves the dialect robustness and leads to promising preliminary steps toward dialect-aware metrics (\S\ref{sec:dialect-awareness}). 
Furthermore, we present evidence that language-aware pretraining can improve the metric performance on the WMT benchmark (\S\ref{sec:wmt-results}) and that the method successfully transfers to other evaluation setups like quality estimation (\S\ref{sec:qe}).

\paragraph{Datasets} As described in \S\ref{sec:method}, we consider micro-level and sentence-level dialect rewrites. The micro-level rewrites are based on pairwise data from \citet{demszky-etal-2021-learning}, in which each example includes a form containing at least one dialect feature from Indian English and a corresponding ``base sentence'' in U.S. English.
We then apply the semantic perturbation to the base sentence as described in \S\ref{sec:micro}. 
For each perturbation type, one of the coauthors manually examined whether the perturbation successfully changes the meaning of the sentence. If all of the three perturbations fail, we exclude the instance from analysis.\footnote{
For the sentences that have multiple dialect rewritings, we treat each one as an individual data point. When multiple semantic perturbations can be applied, we choose a single one, preferring replacements, then insertions, and then deletions.
}

\begin{table}
\small \centering
\begin{tabular}{@{}llll@{}}
\toprule
 & \textbf{EN} & \textbf{PT} & \textbf{ZH} \\ \midrule
All & 148& 2616 & 2227  \\ \midrule
Replace & 96 & 962  & 866 \\
Insert & 89 & 550 & 528  \\
Delete & 63 & 693 & 614  \\ \midrule
\textbf{\textsc{Agg.}} & 115 & 1415 & 1252  \\ \bottomrule
\end{tabular}
\caption{Number of evaluation examples per language before and after  semantic perturbation. The middle three rows are the number of examples to which each perturbation was applicable, and the final row \textsc{Agg.} is the number of examples to which at least one perturbation is applicable, which we use in our final analysis.}
\label{tab:eval_numbers}
\end{table}
\begin{figure*}
    \centering
    \includegraphics[width=\linewidth]{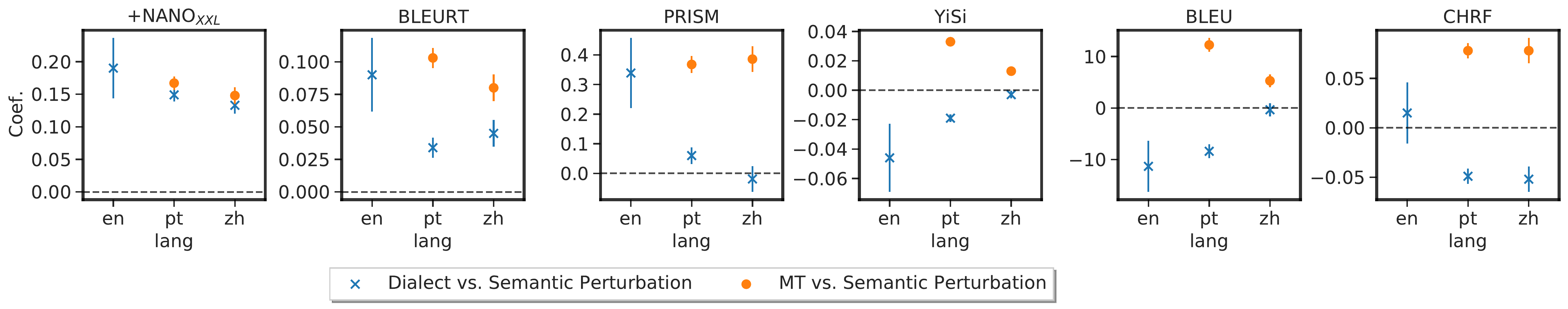}
    \caption{Coefficients from the regression model for \emph{Dialect vs. Semantic Perturbation} ($\phi_{\text{dialect vs. perturb}}$) and \emph{MT vs. Semantic Perturbation} ($\phi_{\text{MT vs. perturb}}$). The higher $\phi_{\text{dialect vs. perturb}}$ is, the more dialect-robust a metric is. Error bars show 99\% confidence intervals; they are larger for the English evaluations because there is less data. $\phi_{\text{MT vs. perturb}}$ serves as a stress test to measure evaluation metrics' abilities of recognizing semantic changes. We show that evaluation metrics are good at recognizing semantic changes but not dialect changes. For all metrics except \bleurt and \sysname, $\phi_{\text{dialect}} - \phi_{\text{perturb}}$ is negative for at least one language, exposing their vulnerability to dialects. }
    \label{fig:coeffs}
\end{figure*}

\begin{table*}[]
\centering\small
\resizebox{\textwidth}{!}{
\begin{tabular}{@{}l|lll|ll||ll|ll|ll@{}}
\toprule
 & \multicolumn{3}{c|}{\textit{Learned}} & \multicolumn{2}{c||}{\textit{Lexical}} & \multicolumn{2}{c|}{\textbf{mT5$_{\text{base}}$}} & \multicolumn{2}{c|}{\textbf{mT5$_{\text{XL}}$}} & \multicolumn{2}{c}{\textbf{mT5$_{\text{XXL}}$}} \\ \midrule
& \textbf{\bleurt} & \textbf{\prism} & \textbf{\yisi} & \textbf{\bleu} & \textbf{\chrf} & \textbf{-\sysname} & \textbf{+\sysname} & \textbf{-\sysname} & \marksymbol{diamond*}{magenta} \textbf{+\sysname} & \textbf{-\sysname} & {\small\faTrophy} \textbf{+\sysname} \\ \midrule
EN &  0.53 & 0.51 & 0.53 & 0.49 & 0.46 & 0.50 & 0.50 & 0.55 & 0.54 & 0.57 & 0.57 \\
 
PT &  \textbf{0.59} & 0.53 & 0.36 & 0.35 & 0.35 & 0.39 & 0.44 & \textbf{0.57} & \textbf{0.65} & \textbf{0.82} & \textbf{0.81}  \\
 
ZH & \textbf{0.59} & 0.47 & 0.46 & 0.35 & 0.36 & 0.46 & 0.45 & 0.51 & \textbf{0.59} & \textbf{0.74} & \textbf{0.74}

 

\\ \bottomrule
\end{tabular}
}
\caption{Success Rates of $\score^{(\text{dialect})} > \score^{(\text{perturb})}$.
Training with \sysname starts to improve upon the strongest baseline BLEURT with mT5$_{\text{XL}}$ (\marksymbol{diamond*}{magenta}) and achieves the best performance with mT5$_{\text{XXL}}$ ({\small\faTrophy}). We \textbf{boldface} the success rates that are better than random chance (0.5) and significant after applying Bonferroni correction for multiple comparisons. Training with \sysname improves dialect robustness for the XL- and base-scale model.}
\label{tab:nano-within}
\end{table*}

For sentence-level dialect analysis, we use the test set of the FRMT benchmark~\citep{FRMT}. Each instance contains an English sentence and its translations into dialects of the target languages Portuguese and Mandarin. For Portuguese, the two dialects are Brazilian Portuguese (pt-BR) and European Portuguese (pt-PT); for Mandarin, we consider mainland  Mandarin  and Taiwanese Mandarin, both in simplified script. As described in \S\ref{sec:sentence-level}, semantic perturbations are obtained by perturbing the English sentences and then translating, using the Google Translate API. \autoref{tab:eval_numbers} shows the number of evaluation examples. 

\subsection{Sensitivity to Dialects}
\label{sec:dialect-robustness}
We use the statistical methods reported in \S\ref{sec:statistical} to test metrics' sensitivity to dialects.

\begin{table}[]
\small\centering
\begin{tabular}{@{}ll|lll@{}}
\toprule
 &  & EN & PT & ZH \\ \midrule
\multirow{2}{*}{mT5$_{\text{base}}$} & -NANO & 0.01$_{0.01}$ & -0.02$_{0.00}$ & -0.02$_{0.00}$ \\
 & +NANO & \textbf{0.04}$_{0.01}$ & \textbf{-0.01} $_{0.00}$ & 0.00$_{0.00}$ \\ \midrule
\multirow{2}{*}{mT5$_{\text{XL}}$} & -NANO & 0.01$_{0.01}$ & 0.02$_{0.00}$ & 0.02$_{0.00}$ \\
 & +NANO & \textbf{0.06}$_{0.01}$ & \textbf{0.05}$_{0.00}$ & \textbf{0.05}$_{0.00}$ \\ \midrule
\multirow{2}{*}{mT5$_{\text{XXL}}$} & -NANO & 0.15$_{0.02}$ & 0.12$_{0.00}$ & 0.11$_{0.00}$ \\
 & +NANO & \textbf{0.19}$_{0.02}$ & \textbf{0.15}$_{0.00}$ & \textbf{0.13}$_{0.00}$ \\ \bottomrule
\end{tabular}
\caption{Coefficients from the regression model for \emph{Dialect vs. Semantic Perturbation}, indicating the dialect robustness, before and after using \sysname. We \textbf{boldface} significant coefficients where \sysname helps.  We show that training with \sysname improves the dialect robustness across all model sizes and languages.}
\label{tab:coefficients-nano}
\end{table}

\paragraph{Regression} Following \autoref{eq:regression}, we use $\score^{(\text{perturb})}_{m, i}$, $\score^{(\text{dialect})}_{m, i}$, $\score^{(\text{MT})}_{m, i}$ as conditions and model each metric as a mixed-effects regression.
For a dialect-robust metric, we expect $\phi_{\text{dialect}} > \phi_{\text{perturb}}$, indicating that dialect rewrites score more highly than semantic perturbations, as required by definition \ref{def:phi-robustness}. The difference $\phi_{\text{dialect}} - \phi_{\text{perturb}}$ is shown in the $Y$-axis of \autoref{fig:coeffs}. We also evaluate $\phi_{\text{MT}} - \phi_{\text{perturb}}$ as a stress test to measure metrics' abilities to recognize semantic changes, and to ensure that the semantic perturbations are effective. For all metrics except \bleurt and \sysname, $\phi_{\text{dialect}} - \phi_{\text{perturb}}$ is negative for at least one language, indicating that these metrics are not dialect robust even in the average case. At the same time, all evaluation metrics can distinguish the \textsc{mt} and \textsc{perturb} conditions, showing that the issue is specific to dialect and not generally applicable to other paraphrases. \autoref{tab:coefficients-nano} shows the coefficients before and after using \sysname, which improves  dialect robustness across all model sizes and languages. 

\paragraph{Success Rates} In \autoref{tab:nano-within} we report the success rates of a metric in assigning higher scores to dialect rewrites than to semantic perturbations.
\bleurt performs better than other existing evaluation metrics which consistently fail to rank the dialect change above the perturbations. However, no metric correctly ranks the English examples at better than a random chance win rate (0.5), and even \bleurt as the most robust metric only has a 0.59 win rate for PT and ZH.
In comparison with \bleurt, \sysname achieves a higher win rate when scaled to XL and XXL, marked with \marksymbol{diamond*}{magenta} in \autoref{tab:nano-within}.
The same trend can be observed in the regression analysis, where \sysname's coefficients are positive for all metrics and languages.
However, the marginal benefit of \sysname over simply finetuning a larger model diminishes at scale---while \sysname leads to significant improvements at XL scale, it has only a minor effect on the XXL-sized model.

\subsection{Dialect Awareness}
\label{sec:dialect-awareness}

Since existing metrics do not train with dialect identifiers, we are only able to test \sysname's performance in terms of dialect awareness following Definition~\ref{def:dialect-awareness}, which can serve as a baseline for future works.

\paragraph{Experiments} We use the Mandarin dataset for sentence-level dialect rewrites for our experiments of dialect awareness, because Mandarin is covered during the pretraining of \sysname.\footnote{We provide the zero-shot result of dialect awareness of \sysname on PT in Appendix~\ref{app:dialect-awareness-pt}.} We then score each dialect rewrite against its translation from the English sentence, written as:
\begin{align}
\score^{j}_{m, i} &= m(\text{tag}, y^{(\text{MT})}_i,  y^{(j)}_i)
\end{align}

The models we use are the ones we trained for dialect robustness tests in \autoref{tab:nano-within}, but we provide specific dialect tags (e.g., zh-CN for Mainland Mandarin) instead of the general language tags (e.g., zh-any) as inputs for inference. During the model inference, we either provide tags that agree or disagree with the candidate sentences. For example, for a candidate sentence in Taiwanese Mandarin, we run inference with both ``zh-CN'' and ``zh-TW''. A dialect-aware metric should assign higher scores for the input with the correct dialect tag.


\begin{table}[]
\small\centering
\resizebox{\columnwidth}{!}{
\begin{tabular}{@{}cc|llll@{}}
\toprule
\textbf{Candidate} & \textbf{Input Tag} & \textbf{-\sysname$_\text{\tiny{XL}}$} & \textbf{+\sysname$_\text{\tiny{XL}}$} & \textbf{-\sysname$_\text{\tiny{XXL}}$} & \textbf{+\sysname$_\text{\tiny{XXL}}$} \\ \midrule
\multirow{2}{*}{zh-TW} 
& zh-TW & 0.70 \xmark  & 0.71 \checkmark & 0.80 \checkmark & 0.75 \xmark \\
& zh-CN & 0.70 & 0.68 & 0.77 & 0.78 \\
\midrule
\multirow{2}{*}{zh-CN} 
& zh-TW & 0.74  & 0.68 & 0.80 & 0.77  \\
& zh-CN & 0.75 \checkmark   & 0.82 \checkmark & 0.76 \xmark & 0.81 \checkmark  \\
 \bottomrule
\end{tabular}
}
\caption{Dialect Awareness test on simplified Mandarin. We score each variant against translation of English to Mandarin, with the dialect tag as input. 
\checkmark{} shows when the metric assign a higher score for the candidate with the matched dialect identifier, indicating the dialect awareness, and \xmark{} shows when it does not. \textsc{NANO}$_\text{\tiny{XL}}$ successfully assigns higher scores when the candidates matches with the input tags.}
\label{tab:awareness_zh}
\end{table}

\paragraph{Results} \autoref{tab:awareness_zh} shows the results of dialect awareness of \sysname. \sysname{}$_\text{\tiny{XL}}$ assigns higher scores to the candidates with the correct dialect tag, compared to the finetuning-only setup ($-\sysname{}{_\text{\tiny{XL}}}$). However, at the XXL scale the picture is more mixed: the \sysname{}$_\text{\tiny{XXL}}$ metric successfully assigns higher scores for zh-CN inputs with zh-CN tag over the zh-TW tag, but it fails on zh-TW inputs. This is compatible with our mixed findings on the impact of \sysname on dialect robustness at the XXL scale.

\subsection{Performance on WMT Tasks}
\label{sec:wmt-results}

We have shown that \sysname is more robust to dialects. Is the robustness at the cost of sacrificing the metrics' performance on standard benchmark of evaluation metrics? To study this, we evaluate on the test set of WMT 2020.

\paragraph{Metrics} We calculate the segment-level agreement with human ratings and report DaRR~\citep{mathur-etal-2020-results}, a robust variant of Kendall Tau. We follow \citet{pu-etal-2021-learning} and omit \texttt{$*$-en} results because of inconsistencies between benchmark implementations.

\paragraph{Results} \autoref{tab:wmt_within} and \autoref{tab:wmt_qe} show the performance of existing methods and \sysname on WMT 2020 test sets for within the same language and quality estimation settings respectively. In both settings, adding \sysname improves mT5$_\text{\tiny{XL}}$ model's performance on WMT benchmark tasks compared to the finetuning-only setup. As in the dialect robustness tests, \sysname does not help much for the model size XXL and achieves comparable results to finetuning-only settings. Moreover, our results are on par with or exceed those of prior metrics, demonstrating that mT5 is an effective base model for developing new metrics.

\subsection{Transfer to Quality Estimation}
\label{sec:qe}
We next adapt \sysname to a quality estimation setting for machine translations following \S\ref{sec:finetune}. The WMT benchmark allows testing metrics in this setting with and without references.

\paragraph{Quality Estimation} While we have been focusing the cross-dialect setting within the same language, all the statistical methods can be applied to the cross-language cross-dialect setting, and training with \sysname can serve as quality estimation of the translation quality. Similar to \S\ref{sec:sentence-level}, given an English base sentence $\en_i$ and its translation to two locales ($j$ and $k$) of a target language. We have
\begin{align}
\score^{j}_{m, i} &= m(\en_i,  y^{(j)}_i) \\
\score^{k}_{m, i} &= m(\en_i,  y^{(k)}_i).
\end{align}

For a system that produces equally-good quality translations that are in different dialects $j$ and $k$, we expect $\mathbb{E}[\score^{j}_{m}] \approx \mathbb{E}[\score^{k}_{m}] > \mathbb{E}[\score^{\text{perturb}}_{m}]$ for a metric that is robust to dialect variations. 

\paragraph{Quality Estimation with Reference} For the quality estimation, we can also use one dialect ($k$) as reference and evaluate other conditions (e.g., perturb, MT, dialect $j$) against dialect $k$ as candidates for evaluation, written as:
\begin{align}
\score^{j}_{m, i} &= m(\en_i,  y^{(j)}_i, y^{(k)}_i) \\
\score^{\text{perturb}}_{m, i} &= m(\en_i, y^{(\text{perturb})}_i, y^{(k)}_i).
\end{align}
For a metric that is robust to dialect variations, we expect $\mathbb{E}[\score^{j}_{m}] > \mathbb{E}[\score^{\text{perturb}}_{m}]$. The candidate can also be $y^{(\text{MT})}_i$. 
We can use all statistical methods in \S\ref{sec:statistical} to understand the difference of outputs from evaluation metrics. 

\begin{figure*}[t]
        \centering
        \subfigure[QE] {
                \includegraphics[width=0.58\textwidth]{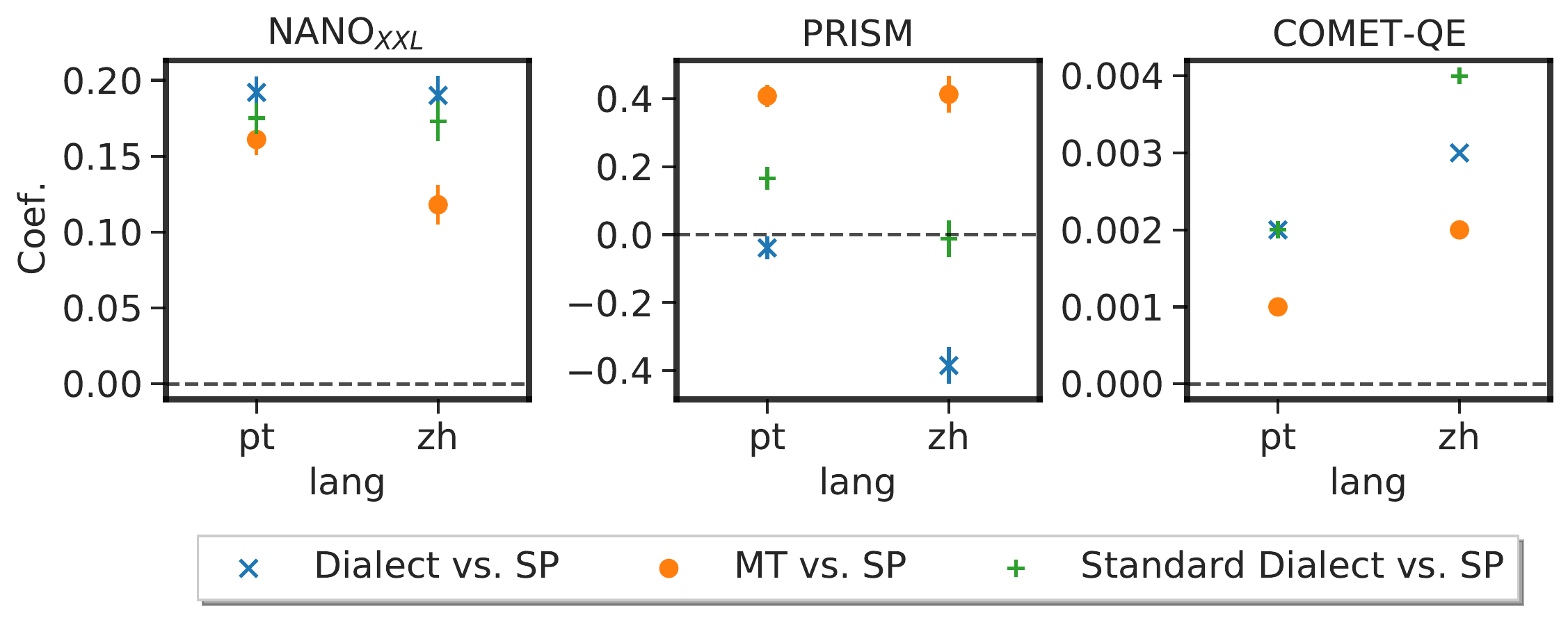}
                \label{fig:qe}
            }
        \subfigure[QE with Reference]  {
                \includegraphics[width=0.385\textwidth]{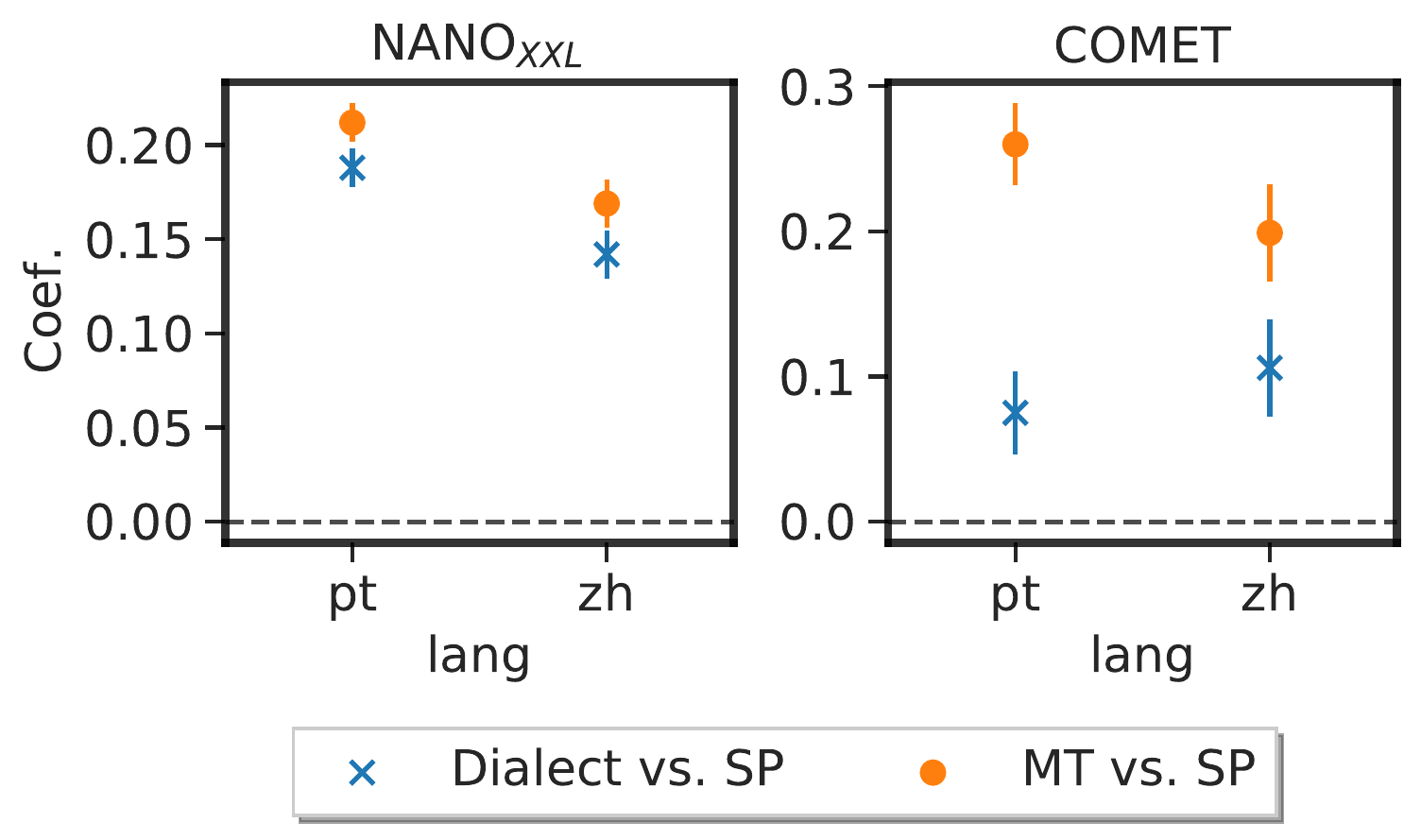}
                \label{fig:qe_w_ref}
        }
        \caption{Coefficients from the regression model for metric as quality estimation without and with references. \sysname consistently rates dialects higher than semantic perturbations for both setups, and assigns similar ratings across dialects (\textcolor{mygreen}{$\bullet$} and \textcolor{myblue}{\ding{53}}) for quality estimation with references.
        }
        \label{fig:qe_coeffs}
\end{figure*}

\begin{table}[]
\resizebox{0.5\textwidth}{!}{
\begin{tabular}{@{}llll|llll@{}}
\toprule
 &  & \textbf{\prism} & \textbf{\comet} & \textbf{-\sysname$_{\text{\tiny{XL}}}$} & \textbf{+\sysname$_{\text{\tiny{XL}}}$}  & 
\textbf{-\sysname$_{\text{\tiny{XXL}}}$} & \textbf{+\sysname$_{\text{\tiny{XXL}}}$} \\ \midrule
\multicolumn{1}{l|}{\multirow{2}{*}{QE}} & PT & 0.44 & 0.54 & 0.67 & 0.76 & 0.84 & 0.85 \\
\multicolumn{1}{l|}{} & ZH & 0.30 & 0.53$^{*}$ & 0.67 & 0.75 & 0.84 & 0.84 \\ \midrule
\multicolumn{1}{l|}{\multirow{2}{*}{\begin{tabular}[c]{@{}l@{}}QE\\$\text{ref}$\end{tabular}}} & PT  & - & 0.53 & 0.63 & 0.64 & 0.86 & 0.85  \\
\multicolumn{1}{l|}{} & ZH & - & 0.53$^{*}$ & 0.55 & 0.55 & 0.79 & 0.75 \\ \bottomrule
\end{tabular}
}
\caption{The success rates of $\score^{(\text{dialect})} > \score^{(\text{perturb})}$ for Quality Estimation without and with references. \sysname improves the dialect robustness upon existing metrics on both quality estimation settings.}
\label{tab:qe}
\end{table}

\paragraph{Experiment Setup} We use the datasets for sentence-level dialect features for the quality estimation with and without references experiments. For quality estimation, we take the English sentences as the source and candidate from each of four conditions: two human-written dialects of target language (e.g., pt-BR), translated outputs to target language from English and semantic perturbation as the input for the quality estimation. The translated outputs are from the Google Translate API. If a metric is robust to dialects, we expect $\mathbb{E}[\score^{\text{dialect}}_{m}] \geq \mathbb{E}[\score^{\text{MT}}_{m}] > \mathbb{E}[\score^{\text{perturb}}_{m}]$.  For quality estimation with reference, we keep the same setting as the quality estimation but use one of the two dialects (``zh-CN'' for Mandarin and ``pt-BR'' for Portuguese) as reference. We then use \{perturb, MT, the other dialect\} as candidates and estimate their quality with regard to the selected dialects.

\begin{table}[]
  \small\centering
  \resizebox{\columnwidth}{!}{
    \begin{tabular}{@{}l|l|lllllll@{}}
        \toprule
        & \textbf{en-*} & \textbf{en-cs} & \textbf{en-de} & \textbf{en-ja} & \textbf{en-pl} & \textbf{en-ru} & \textbf{en-ta} & \textbf{en-zh} \\ \midrule
        BLEURT & 55.2 & 70.8 & 45.3 & 63.0 & 51.0 & 36.8 & 67.9 & 51.6 \\
        Prism & - & 63.8 & 39.8 & 60.2 & 46.0 & 33.9 & - & 41.6 \\
        YiSi & 35.6 & 50.1 & 32.7 & 44.8 & 21.7 & 24.0 & 35.7 & 40.0 \\ 
        \midrule\midrule
        -\sysname$_\text{\tiny{XL}}$ & 49.2 & 68.2 & 41.0 & 63.0 & 48.6 & 30.8 & 68.5 & 51.0 \\
        +\sysname$_\text{\tiny{XL}}$ & 54.2 & 69.8 & 41.9 & 63.7 & 49.9 & 33.2 & 70.2 & 50.9 \\
        \midrule
        -\sysname$_\text{\tiny{XXL}}$ & 58.6 & 73.0 & 47.9 & 66.3 & 54.1 & 38.7 & 72.0 & 58.1 \\
        +\sysname$_\text{\tiny{XXL}}$ & 58.3 & 72.4 & 47.1 & 66.3 & 53.6 & 39.4 & 72.2 & 56.9 \\ \bottomrule
    \end{tabular}
    }        
    \caption{Segment-level agreement with human ratings on the WMT 2020 test set. The metric is WMT Metrics DaRR~\citep{mathur-etal-2020-results}, a robust variant of Kendall Tau and higher is better.}
    \label{tab:wmt_within}
\end{table}

\paragraph{Results} We show that success rates of $\score^{(\text{dialect})} > \score^{(\text{perturb})}$ in QE with and without references in \autoref{tab:qe}. The setting is pretraining with all data. 
We show that 1) \sysname outperforms existing metrics on dialect robustness for both Portuguese and Mandarin; 2) pretraining stage of \sysname is important to improve dialect robustness with a smaller model size (i.e., mT5$_\text{XL}$ in our case). The trends are consistent with our findings for the within-language evaluation. \autoref{fig:qe_coeffs} shows the coefficients from the regression model and confirms the dialect robustness of \sysname by assigning higher scores to dialects than semantic perturbations.

\begin{table}[]
  \small\centering
  \resizebox{\columnwidth}{!}{
    \begin{tabular}{@{}l|l|lllllll@{}}
        \toprule
         & \textbf{en-*} & \textbf{en-cs} & \textbf{en-de} & \textbf{en-ja} & \textbf{en-pl} & \textbf{en-ru} & \textbf{en-ta} & \textbf{en-zh} \\ \midrule
        COMET & 51.4 & 70.9 & 37.3 & 51.5 & 48.9 & 39.4 & 61.3 & 50.3 \\
        Prism  & - & 48.3 & 26.5 & 38.2 & 18.8 & 11.6 & - & 11.3 \\ 
        \midrule\midrule
        -\sysname$_\text{\tiny{XL}}$ & 51.4 & 68.7 & 40.6 & 59.6 & 44.3 & 28.2 & 66.3 & 51.8 \\
        +\sysname$_\text{\tiny{XL}}$ & 53.8 & 69.5 & 42.7 & 62.6 & 47.1 & 31.5 & 68.4 & 54.8 \\
        \midrule
        -\sysname$_\text{\tiny{XXL}}$ & 57.4 & 71.4 & 47.1 & 65.5 & 52.4 & 36.3 & 70.3 & 58.7   \\
        +\sysname$_\text{\tiny{XXL}}$ & 57.6 & 71.8 & 46.6 & 66.3 & 51.0 & 38.5 & 70.4 & 58.8 \\ \bottomrule
    \end{tabular}
    }
    \caption{Segment-level agreement with human ratings for metrics as quality estimation without references. }
    \label{tab:wmt_qe}
\end{table}

\section{Related Work}
Most popular NLP datasets and evaluation metrics do not take dialectal variation into consideration. For example, machine translation systems are usually evaluated by whether they match references in the target language, for which the dialect is generally unspecified~\citep{DBLP:journals/corr/abs-2202-06935}. 
The subcommunity that has attended most to dialect is the VarDial series of workshops, which has featured shared tasks such as dialect classification~\citep{zampieri2014report}, translation between dialects~\citep{akhbardeh-etal-2021-findings}, and transfer of NLP systems across dialects ~\citep{zampieri2017findings}. Of this prior work, dialect classification is clearly relevant to the criterion of dialect awareness introduced in Definition~\ref{def:dialect-awareness}~\citep[see also][]{gabmap}, but our goal is to reward system outputs that match a target dialect rather than to classify the dialect of existing human-written texts. 
A related line of work has focused on inducing dialect features from corpora~\citep{Eisenstein2010ALV,jorgensen-etal-2015-challenges,DBLP:journals/corr/abs-2104-01299} and on recognizing dialect features in text~\citep{demszky-etal-2021-learning,masis2022corpus}. Following the feature-based view of dialect, we use cross-dialectal minimal pairs to measure dialect robustness in \S\ref{sec:micro}.

On the more specific topic of dialect-aware evaluation, classic approaches focused on the creation of dialect-specific test sets, e.g.~for translation to and from Arabic dialects~\citep[e.g.,][]{zbib-etal-2012-machine}. This idea has been extended to modern multi-task natural language understanding benchmarks by the VALUE project~\citep{value}, which used transformation rules to convert the GLUE benchmarks~\citep{wang2018glue} 
into African-American English. 
Our evaluation in \S\ref{sec:sentence-level} builds on the FRMT dataset of multi-dialectal translations~\citep{FRMT} to evaluate metrics for dialect robustness. However, in many cases it is not feasible to produce multi-dialect references or test sets. In these cases, dialect-robust and dialect-aware metrics can provide a more efficient solution, particularly if these capabilities can be achieved through a pretraining step like \sysname, which can be transferred to multiple tasks and evaluation settings.

Our work is additionally motivated by several papers that demonstrate the social impact of the failure to consider dialect variation in language technology. For example, literature shows that the out-of-the-box POS taggers \citep{jorgensen-etal-2015-challenges} and language identification and dependency parsing tools \citep{blodgett-etal-2016-demographic} perform poorly on AAVE texts.
Other work has demonstrated large racial disparities in the performance of commercial speech recognition systems~\citep{DiChristofano2022PerformanceDB,koenecke2020racial}. \citet{sap-etal-2019-risk} show that models for the detection of toxicity have significantly more false positives for African American English. Our results contribute to this line of work by showing that metrics for text generation tend to penalize dialectal variants. We view the design of dialect-robust and dialect-aware metrics like \sysname{} as a step towards making language technology that works more broadly across dialects.

\section{Conclusion and Future Work}
\label{sec:conclusion}
We introduce and formalize the dialect robustness and dialect awareness in the context of generation evaluation. Grounded by a suite of statistical tests, we find that existing evaluation methods are not robust to dialects. As a first step toward a solution to this problem, we propose \sysname as a pretraining strategy. Our experiments demonstrate that \sysname offers a size-efficient way to improve both the dialect robustness, shows the preliminary success towards dialect awareness and improves the metric performance of metrics on WMT benchmark. 

Due to the limited availability of dialect-parallel corpora, our robustness tests are conducted in thousands of examples for Mandarin and Portuguese and hundreds of examples for English, which is insufficient to capture the full extent of these languages. We encourage future work to develop more resources, including benchmarks and corpora to conduct research on dialects for NLG evaluation. 
Due to this limitation, our work focuses on dialect robustness and only briefly evaluates dialect awareness. Future works may extend the details and criteria of the dialect-aware NLG evaluation, and we hope our work can serve as a baseline in this new research direction. 
Our encouraging preliminary results lead us to urge researchers to consider and improve the dialect diversity during pretraining.

\section*{Limitations}
Besides the limitations of corpora size for evaluation and a brief exploration of dialect awareness that we state in \S\ref{sec:conclusion}, we again acknowledge the data acquisition strategy as another limitation of our work. Our data acquisition of dialects requires country codes, which exclude many dialects. 
There is some work on getting dialectal data without country codes: \citet{blodgett-etal-2016-demographic} build a dataset of tweets that are likely to include a high density of African-American English by linking geolocated Twitter data with demographic data from the U.S. census. However, this approach is limited to dialects that have strong geographic associations within the United States and which correlate with census demographics like race.
Similarly, \citet{abdul-mageed-etal-2018-tweet} build a dataset of city-level Arabic dialects, again relying on Twitter geolocation.
An alternative approach that does not rely on geolocation is to translate existing corpora into multiple dialects~\citep[e.g.,][]{faisal-etal-2021-sd-qa,value}. However, this is labor intensive and therefore difficult to scale up to the amount of data needed for pretraining. We leave to future work the question of how to build large-scale corpora for dialects that do not align with easily-identifiable geographical indicators such as national boundaries.
\section*{Acknowledgement}
We would like to thank Dan Deutsch and Jonathan Clark for their precious feedback that helps improve the draft. We also thank Jason Riesa, Markus Freitag, Xavier Garcia, Qijun Tan and Nanyun Peng for their insightful discussion.  The technical implementation of \sysname has benefited from the support of Google T5X team and Xavier Garcia.

\newpage
\bibliography{main}
\bibliographystyle{acl_natbib}

\newpage\clearpage
\appendix
\begin{table*}[]
\small\centering
\begin{tabular}{@{}l|l|l|l@{}}
\toprule
 & \textbf{Task Instruction} & \textbf{Examples} & \textbf{Output Prefix} \\ \midrule\midrule
\multirow{6}{*}{Delete} & \multirow{6}{*}{\begin{tabular}[c]{@{}l@{}}Generate a \\ sentence by \\ deleting one\\ word from \\ the original \\ sentence and \\ change its \\ meaning.\end{tabular}} & \textbf{Original Sentence:} & \textbf{\textbackslash{}nDelete one word from original sentence:} \\ \cmidrule(l){3-4} 
 &  & \begin{tabular}[c]{@{}l@{}}the person I like \inlineyellowbox{the most} is from the \\ mechanical department\end{tabular} & \begin{tabular}[c]{@{}l@{}}the person I like is from the \\ mechanical department\end{tabular} \\ \cmidrule(l){3-4} 
 &  & \inlineyellowbox{a recipe} is a simple thing & it is a simple thing \\ \cmidrule(l){3-4} 
 &  & \begin{tabular}[c]{@{}l@{}}the union person contacted his \\ representative \inlineyellowbox{at the school}\end{tabular} & \begin{tabular}[c]{@{}l@{}}the union person contacted his\\ representative\end{tabular} \\ \cmidrule(l){3-4} 
 &  & \begin{tabular}[c]{@{}l@{}}we have two tailors who can make \\ them \inlineyellowbox{for us}\end{tabular} & \begin{tabular}[c]{@{}l@{}}we have two tailors who can make \\ them\end{tabular} \\ \cmidrule(l){3-4} 
 &  & \begin{tabular}[c]{@{}l@{}}So if you're \inlineyellowbox{not} good at communication\\  you may get filtered at even the first level\end{tabular} & \begin{tabular}[c]{@{}l@{}}So if you're good at communication\\ you may get filtered at even the first level\end{tabular} \\ \midrule
\multirow{6}{*}{Replace} & \multirow{6}{*}{\begin{tabular}[c]{@{}l@{}}Generate a \\ sentence by \\ replacing \\ one word \\ from the \\ original \\ sentence and \\ change its \\ meaning.\end{tabular}} & \textbf{Original Sentence:} & \textbf{\textbackslash{}nReplace one word from original sentence:} \\ \cmidrule(l){3-4} 
 &  & \begin{tabular}[c]{@{}l@{}}the person I like the \inlinebluebox{most} is from the \\ mechanical department\end{tabular} & \begin{tabular}[c]{@{}l@{}}the person I like the least is from the \\ mechanical department\end{tabular} \\ \cmidrule(l){3-4} 
 &  & a recipe is a \inlinebluebox{simple} thing & a recipe is a complicated thing \\ \cmidrule(l){3-4} 
 &  & \begin{tabular}[c]{@{}l@{}}the union person contacted his \\ representative at the \inlinebluebox{school}\end{tabular} & \begin{tabular}[c]{@{}l@{}}the union person contacted his \\ representative at the factory\end{tabular} \\ \cmidrule(l){3-4} 
 &  & \begin{tabular}[c]{@{}l@{}}we have \inlinebluebox{two} tailors who can make \\ them for us\end{tabular} & \begin{tabular}[c]{@{}l@{}}we have three tailors who can make \\ them for us\end{tabular} \\ \cmidrule(l){3-4} 
 &  & he didn't give it to \inlinebluebox{me} & he didn't give it to anyone \\ \midrule
\multirow{6}{*}{Insert} & \multirow{6}{*}{\begin{tabular}[c]{@{}l@{}}Add one \\ word to \\ a sentence \\ and change\\ its meaning.\end{tabular}} & \textbf{Original Sentence:} & \textbf{\textbackslash{}nAdd a word to it:} \\ \cmidrule(l){3-4} 
 &  & it was the first day of term & it was the first day of \inlinepurplebox{spring} term \\ \cmidrule(l){3-4} 
 &  & \begin{tabular}[c]{@{}l@{}}the person I like the \\ most is from the mechanical department\end{tabular} & \begin{tabular}[c]{@{}l@{}}the person I like \inlinepurplebox{to talk to} the \\ most is from the mechanical department\end{tabular} \\ \cmidrule(l){3-4} 
 &  & he does a lot of things & he does a lot of \inlinepurplebox{funny} things \\ \cmidrule(l){3-4} 
 &  & \begin{tabular}[c]{@{}l@{}}my brother said that one of his favorite \\ places is the beach nearby\end{tabular} & \begin{tabular}[c]{@{}l@{}}my brother said that one of his \inlinepurplebox{least} favorite\\ places is the beach nearby\end{tabular} \\ \cmidrule(l){3-4} 
 &  & \begin{tabular}[c]{@{}l@{}}I think you should start going to the\\ gym from now on\end{tabular} & \begin{tabular}[c]{@{}l@{}}I think you should start going to the\\\inlinepurplebox{other} gym from now on\end{tabular} \\ \bottomrule
\end{tabular}
\caption{The prompts, prefix and five examples that we use to prompt LaMDA for automatic semantic perturbation on English sentences. We include three types of semantic perturbation: replace (highlighted in \inlineyellowbox{yellow}), delete (highlighted in \inlinebluebox{blue}) and insert (highlighted in \inlinepurplebox{purple}). }
\label{tab:semantic_perturb}
\end{table*}

\section{Five Examples for Semantic Perturbation}
\label{app:five_shots}

\autoref{tab:semantic_perturb} shows the task instruction and examples we used to prompt LaMDA for the automatic semantic perturbation on English sentences, for both micro-level (\S\ref{sec:micro}) and sentence-level (\S\ref{sec:sentence-level}) studies. During decoding, we use greedy decoding.
\section{Languages and Variants}
\label{app:lang_variants}

\begin{table*}[]
\small
\centering
\begin{tabular}{@{}l|l@{}}
\toprule
\textbf{Language} & \textbf{Region Code} \\ \midrule
en & \begin{tabular}[c]{@{}l@{}}AU, BZ, BM, BR, CA, KY, CK, CU, DO, FK, GI, GP, GT, GY, HN, IE, LR, MX, NF, \\ PN, SH, ZA, SR, \underline{GB}, \underline{US}, VE, \underline{IN}\end{tabular} \\ \midrule
cs & AT, CZ, PL, SK \\ \midrule
de & AT, BE, CZ, DK, FR, DE, HU, IT, LI, LU, NL, PL, RO, SK, SI, CH \\ \midrule
ja & JP \\ \midrule
km & KH, LA, TH, VN \\ \midrule
pl & BY, CZ, DE, LT, PL, RU, SK, UA \\ \midrule
ps & PK \\ \midrule
ru & BY, CN, EE, FI, GE, KZ, KP, KG, LV, LT, MD, MN, NO, PL, RO, RU, TM, UA, UZ \\ \midrule
ta & IN, LK \\ \midrule
zh-cmn-Hans & CN, KP, LA, VN, TW, MM, MN, RU \\ \midrule
zh-yue & CN, VN, HK \\ \midrule
zh-cmn-Hant & CN, TW \\ \bottomrule
\end{tabular}
\caption{The language code and region code that we cover. We consider 10 WMT languages and use BCP language codes. We \underline{underline} selected English dialects under the increasing noise setup \texttt{zh, pt, en} in \S\ref{sec:pretraining}.}
\label{tab:langs}
\end{table*}

\autoref{tab:langs} shows the language codes and region codes that we cover during \sysname~pretraining. We cover 10 WMT languages and 95 language variants, presented as BCP language codes. Although \texttt{iu} is one of the WMT languages, it is not supported by LangID model that we are using and we thus do not include it in our pretraining. Portuguese (PT) is not a WMT language and therefore not included. Therefore, all \sysname~dialect robustness results on PT are fully through zero-shot transfer. We report additional experiments that include Portuguese during pretraining in Appendix~\ref{app:ablation}. Our experiments show that pretraining with all languages leads to better dialect robustness on both PT and ZH.

\section{Metric Implementations}
\begin{table}[]
\small\centering
\resizebox{0.5\textwidth}{!}{
\begin{tabular}{@{}c|ccccc@{}}
\toprule
 & BLEURT & PRISM & YiSi & COMET & NANO \\ \midrule
Within & $\checkmark$ & $\checkmark$ & $\checkmark$ &  & $\checkmark$ \\
QE &  & $\checkmark$ &  & $\checkmark$ &  $\checkmark$\\
QE w/ Ref &  &  &  & $\checkmark$ & $\checkmark$ \\ \bottomrule
\end{tabular}
}
\caption{Supported setups for different metrics.}
\vspace{-0.2cm}
\label{tab:capability}
\end{table}

We use the official releases of Prism~\cite{prism}, COMET~\cite{comet} and BLEURT~\cite{pu-etal-2021-learning} in our work. For YiSi, we use an internal implementation. \autoref{tab:capability} presents the supported setups for each model in their latest released versions. Although all metrics could in theory be adapted to different use cases, their existing capabilities restrict the experiments we can run with them. For BLEURT,\footnote{\url{https://github.com/google-research/bleurt}.} we use the latest checkpoint \texttt{BLEURT-20}. We use Prism\footnote{\url{https://github.com/thompsonb/prism}.} (\texttt{m39v1} checkpoint) for quality estimation with and without references. Lastly, there are two models that we use for COMET\footnote{\url{https://github.com/Unbabel/COMET}}. Model \texttt{wmt21-comet-qe-mqm} is for reference-free quality estimation and \texttt{wmt20-comet-da} for reference-based quality estimation. For our experiments, if a language is not supported by the model. we will exclude it from the results.
\section{Training Details and Hyperparameters of \sysname}
\label{app:training}
\paragraph{Hyperparameters} We implement \sysname using T5X and SeqIO~\cite{roberts2022t5x}. We experimented with the following hyperparemeters during training: learning rate of \{$1e-3, 1e-4, 1e-5, 3e-5, 5e-5$\} $\times$ sequence length of \{512, 1024\}.  The reported results are based on a learning rate of $1e-4$ and sequence length of 1024. We train for 200,000 steps for pretraining and another 20,000 steps for finetuning. We set the drop out rate as 0.1 and optimizer warm up steps as 10,000. We train with a batch size of 128. 

\paragraph{Choosing Checkpoints} We calculate the Kendall-Tau correlation on the development set every 1000 steps throughout training and choose the checkpoint with the highest correlation as the final checkpoint for evaluation.

\paragraph{Compute Time} Our models are trained on 64 TPUs, pretraining step normally takes one day to finish across different sizes. While mT5$_{\text{small}}$ can be trained within a single day, finetuning mT5$_{\text{XL}}$ and mT5$_{\text{XXL}}$ takes three and nine days respectively to reach 20,000 steps, but the models converge before they finish training.

\section{\sysname Design Choices}
\label{app:ablation}

\begin{figure*}
    \centering
    \includegraphics[width=\linewidth]{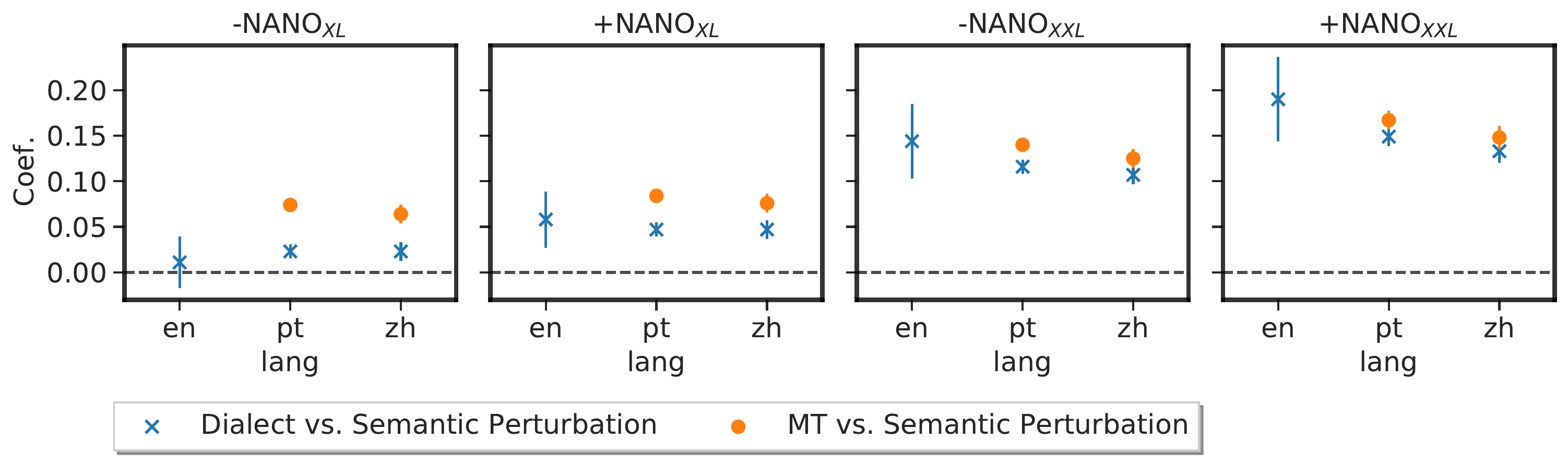}
    \vspace{-0.5cm}
    \caption{Coefficients from the regression model for \emph{Dialect vs. Semantic Perturbation} ($\phi_{\text{dialect vs. perturb}}$) and \emph{MT vs. Semantic Perturbation} of \sysname across XL and XXL model sizes. Training with \sysname improves the dialect robustness for both model sizes. This figure is complementary to \autoref{fig:coeffs}.}
    \label{fig:nano_scales}
    \vspace{-0.3cm}
\end{figure*}

\begin{table*}[]
\small\centering
\resizebox{\textwidth}{!}{
\begin{tabular}{@{}ll|ll|llll|llll@{}}
\toprule

 & & \multicolumn{2}{c|}{EN} & \multicolumn{4}{c|}{PT} & \multicolumn{4}{c}{ZH} \\ \cmidrule{2-12}
 & & $\phi_{\text{dl vs. pb}}\uparrow$ & \underline{$R_{\text{pb}}$}$\uparrow$ & $\phi_{\text{dl vs. pb}}\uparrow$ & $\phi_{\text{dl vs. MT}}$ &  \underline{$R_{\text{pb}}$}$\uparrow$  & \underline{$R_{\text{MT}}$}  & $\phi_{\text{dl vs. pb}}\uparrow$ & $\phi_{\text{dl vs. MT}}$ &  \underline{$R_{\text{pb}}$}$\uparrow$  & \underline{$R_{\text{MT}}$} \\ \cmidrule{2-12}
& BLEURT & 0.09$_{0.01}$  & 0.53$^{*\dagger}$  & 0.03$_{0.01}$ & -0.07$_{0.01}$ & 0.59 & 0.19 & 0.04$_{0.01}$ & -0.04$_{0.01}$ & 0.59  &  0.33 \\ \midrule

\multirow{2}{*}{mT5$_\text{base}$} & Finetuning & 0.01$^{*}_{0.01}$ & 0.50$^{*\dagger}$ & -0.02$_{0.00}$ & -0.09$_{0.00}$ & 0.39 & 0.13 & -0.02$_{0.00}$ & -0.08$_{0.00}$ & 0.46$^{\dagger}$ & 0.31 \\
& \sysname$_{\text{all | }\lambda=1}$ & 0.04$_{0.01}$ & 0.50$^{*\dagger}$ & -0.01$_{0.00}$ & -0.08$_{0.00}$ & 0.44 & 0.16 & 0.00$_{0.00}^{*}$ & -0.08$_{0.00}$ & 0.45$^{\dagger}$ & 0.28 \\
\midrule

\multirow{4}{*}{mT5$_\text{XL}$} &  Finetuning & 0.01$^{*}_{0.01}$ & 0.55$^{*\dagger}$ & 0.02$_{0.00}$ & -0.05$_{0.00}$ & 0.57 & 0.21 & 0.02$_{0.00}$& -0.04$_{0.00}$ & 0.51$^{*}$ & 0.31 \\
& \marksymbol{diamond*}{magenta} \sysname$_{\text{all | }\lambda=1}$ & 0.06$_{0.01}$ & \textbf{0.54}$^{*\dagger}$ & 0.05$_{0.00}$ & -0.04$_{0.00}$ & \textbf{0.65} & \textbf{0.25} & 0.05$_{0.00}$ & -0.03$_{0.00}$ & \textbf{0.59} & \textbf{0.35} \\
\cmidrule{2-12}

& \sysname$_{\text{zh/pt | }\lambda=1}$ & 0.03$_{0.01}$ & 0.53$^{*\dagger}$ & 0.03$_{0.00}$ & -0.04$_{0.00}$ & 0.59 & 0.23  & 0.03$_{0.00}$ & -0.03$_{0.00}$ & 0.54$^{\dagger}$ & 0.32  \\
& \sysname$_{\text{zh/pt/en | }\lambda=1}$ & 0.06$_{0.01}$ & 0.53$^{*\dagger}$ & 0.04$_{0.00}$ & -0.04$_{0.00}$ & 0.64 & 0.24 & 0.04$_{0.00}$ & -0.03$_{0.00}$ & 0.57 & 0.33 \\
\cmidrule{2-12}
& \sysname$_{\text{all | }\text{pos:neg}=2}$ & 0.21$_{0.02}$ & 0.53$^{*\dagger}$ & 0.04$_{0.00}$ & -0.04$_{0.00}$ & 0.60 & 0.23 & 0.04$_{0.00}$ & -0.03$_{0.00}$ & 0.56 & 0.33 \\
\midrule

 \multirow{6}{*}{mT5$_\text{XXL}$} &  Finetuning & 0.15$_{0.02}$ & 0.57$^{*\dagger}$ & 0.12$_{0.00}$ & -0.02$_{0.00}$ & 0.82 & 0.32 & 0.11$_{0.00}$ & -0.02$_{0.00}$ & 0.74 & 0.38 \\
&{\small\faTrophy} \sysname$_{\text{all | }\lambda=1}$ & 0.19$_{0.02}$ & 
\textbf{0.57$^{*\dagger}$} & 0.15$_{0.00}$ & -0.02$_{0.02}$ & \textbf{0.81} & \textbf{0.35} & 0.13$_{0.00}$ & -0.01$_{0.00}$ & \textbf{0.74} & \textbf{0.38} \\

\cmidrule{2-12}
& \sysname$_{\text{zh/pt | }\lambda=1}$ & 0.19$_{0.02}$ & 0.54$^{*\dagger}$ & 0.13$_{0.00}$ & -0.02$_{0.00}$ & 0.80 & 0.33 & 0.12$_{0.00}$ & -0.01$_{0.00}$ &  0.73 & 0.41 \\
& \sysname$_{\text{zh/pt/en} | \lambda=1}$ & -0.18$_{0.02}$ & 0.56$^{*\dagger}$ & 0.13$_{0.00}$ &  -0.02$_{0.00}$ & 0.80 & 0.34 & 0.12$_{0.00}$ & -0.02$_{0.00}$ & 0.73 & 0.39 \\

\cmidrule{2-12}
& \sysname$_{\text{all | }\lambda=0}$ & 0.20$_{0.02}$ & 0.53$^{*\dagger}$ & 0.15$_{0.00}$ & -0.02$_{0.02}$ & 0.82 & 0.35 & 0.13$_{0.00}$ & -0.01$_{0.02}$ & 0.76 & 0.40 \\
& \sysname$_{\text{all | }\lambda=2}$ & 0.20$_{0.02}$ & 0.56$^{*\dagger}$  & 0.15$_{0.00}$ & -0.02$_{0.02}$ & 0.81 & 0.34 & 0.13$_{0.00}$ & -0.01$_{0.00}$ & 0.75 & 0.40 \\ 
\bottomrule
\end{tabular}
}
\vspace{-0.1in}
\caption{Dialect Robustness Tests for metrics with and without \sysname. ``pb'' and ``dl'' are short for ``perturb'' and ``dialect''. $R_{\text{pb}}$ and $R_{\text{MT}}$ are the success rates of $\score^{(\text{dialect})}>\score^{(\text{perturb})}$ and $\score^{(\text{dialect})}>\score^{(\text{MT})}$ correspondingly. Standard errors of coefficients are in the subscript. We can observe that that 1) pretraining improves the dialect robustness compared to the finetuning-only setting and 2) Pretraining on more languages improves the dialect robustness. $\uparrow$ means higher is better. The success rates (\underline{$R$}) are comparable across metrics, but co-efficients from regression models are only comparable within the same metric. \sysname based on mT5$_{\text{XL}}$ with full data improves upon the strongest baseline \marksymbol{diamond*}{magenta} and achieves the best performance with mT5$_{\text{XXL}}$. This is complementary to \autoref{tab:nano-within}.
}
\vspace{-0.4cm}
\label{tab:nano-within-detail}
\end{table*}

Table~\ref{tab:nano-within-detail} shows different variations of \sysname and their performances. We studied:
\begin{itemize}[leftmargin=*]
\itemsep-.3em 
\vspace{-0.2em}
    \item Comparing pretraining on all WMT language variants to only prertaining on zh/pt or zh/pt/en.
    \item Comparing $\lambda=1$ to $\lambda=0$ and $\lambda=2$, i.e., the balance in pretraining between dialect-tags and language-tags.
    \item Variations of the ratio of positive vs. negative instances during pretraining. We compare a balanced set to a setup where we have twice as many positive as negative examples.
\end{itemize}

\noindent We gain the following insights: 1) using all WMT languages for pretraining performs better than using partial data; 2) An equal balance between dialect-tags and general language tags ($\lambda=1$) during pretraining improves upon a higher fraction of dialect-tags ($\lambda=2$). However, using \textit{only} data with general language tags ($\lambda=0$) surprisingly leads to an even better \textbf{dialect-robustness}, although the model will lose its potential for \textbf{dialect-awareness} since it never sees dialect tags; 3) A balanced set of positive and negative instances during pretraining is better than oversampling positive instances.

\begin{table}[]
\centering\small
\resizebox{0.5\textwidth}{!}{
\begin{tabular}{@{}l|l|lll|ll@{}}
\toprule
\textbf{} & \textbf{} & \textbf{\textbf{BLEURT}} & \textbf{PRISM} & \textbf{YiSi} & \textbf{BLEU} & \textbf{CHRF} \\ \midrule
 & EN & 0.10$_{0.01}$ & 0.34$_{0.05}$ & -0.05$_{0.01}$ & -12.01$_{1.91}$ & 0.03$^{*}_{0.01}$ \\
 & PT & 0.03$_{0.00}$ & 0.06$_{0.01}$ & -0.02$_{0.00}$ & -8.39$_{0.53}$ & -0.05$_{0.00}$ \\
\multirow{-3}{*}{\begin{tabular}[c]{@{}l@{}} $\phi_{\text{dl vs. pb}}$ \end{tabular}} & ZH & 0.04$_{0.00}$& -0.02$^{*}_{0.02}$ & -0.00$_{0.00}$ & -0.34$^{*}_{0.49}$ & -0.05$_{0.00}$ \\ \midrule
\bottomrule
\end{tabular}
}
\caption{ Coefficients from Equation~\ref{eq:regression} with standard errors in the subscript. We mark the ones that have $p$-value $\geq$ 0.05/5 = 0.01 with * using Bonferroni correction per metric. $\phi_{\text{dl vs. pb}}$ indicates the corresponding score increase (positive value) for dialect (dl) edits compared to the semantic perturbation (pb). 
For a dialect-robust metric, we expect $\phi_{\text{dl vs. pb}}$ to be positive.}
\label{tab:regression}
\end{table}

Following \autoref{eq:regression}, we use $\score^{(\text{perturb})}_{m, i}$, $\score^{(\text{dialect})}_{m, i}$, $\score^{(\text{MT})}_{m, i}$ as conditions and model each metric as a mixed-effects regression. \autoref{tab:regression} shows $\phi_{\text{dialect}}$ 
with its standard errors against the $\phi_{\text{perturb}}$ condition. Take $\phi_{\text{perturb}}$ for BLEURT under EN as an example, -0.09 with an error smaller than 0.05 means that semantic perturbation would result in a decrease of 0.09 point for BLEURT compared to the dialect condition, and the result is significant. For a dialect-robust metric, we expect its $\phi_{\text{perturb}}$ to be positive. However, this is not always true during our observations. BLEURT performs the best among existing evaluation metrics and all other existing metrics have positive $\phi_{\text{perturb}}$ for at least one language of our test data. This indicates that existing evaluation metrics wrongly assign a higher score to semantically-perturbed rewriting than the dialects in at least one of the three languages, suggesting that they should not be used to assess dialects they were not trained for.
\section{Versatility of \sysname}
\subsection{Input Format}
\label{app:input_format}
We use the following input format to adapt \sysname to different use cases.
\begin{itemize}[leftmargin=*]
\itemsep-.3em 
\vspace{-0.2em}
    \item For within-language assessment, we format the input as  \texttt{candidate}: \emph{\{sentence\}} \texttt{reference}: \emph{\{reference\}}  \texttt{language}: \emph{\{language\_tag\}}.
    \item For quality estimation without reference, we format the input as \texttt{candidate}: \emph{\{sentence\}} \texttt{source}: \emph{\{source\}}  \texttt{language}: \emph{\{language\_tag\}}.
    \item For quality estimation with reference, we format the input as \texttt{candidate}:  \emph{\{sentence\}} \texttt{reference}: \emph{\{reference\}} \texttt{source}: \emph{\{source\}}  \texttt{language}: \emph{\{language\_tag\}}.
\end{itemize}
The \emph{\{language\_tag\}} during fine-tuning indicates the language where the candidate sentence comes from, but it is the general language tag (e.g., ``en-any'') and does not contain the dialect information. We finetune one model for each setting.
\subsection{Dialect Robustness}
We show additional results of coefficients from the regression model across XL and XXL sizes in \autoref{fig:coeffs}, which shows that training with \sysname improves the dialect robustness across both sizes and for all languages. \autoref{tab:nano-within-detail} shows the exact numbers for both coefficients and success rates. \sysname improves the success rates under the XL size, but reach comparable results with training without \sysname under the XXL size. We suspect the discrepancy between getting a higher coefficients but having nearly the same success rates is because some big increase of score after applying \sysname which does not influence the success rates.

\subsection{Performance on Reference-based QE}
\begin{table}[]
\resizebox{0.5\textwidth}{!}{
\begin{tabular}{@{}l|l|lllll@{}}
\toprule
 &   & COMET & FT$_{\text{XL}}$ & \sysname$_{\text{XL}}$  & FT$_{\text{XXL}}$ & \sysname$_{\text{XXL}}$ \\ \midrule
\multicolumn{1}{l|}{\multirow{2}{*}{PT}} & R$_{\text{pb}}$ & 0.54 & 0.67 & 0.76 & 0.84 & 0.85 \\
\multicolumn{1}{l|}{} & R$_{\text{MT}}$ & 0.52 & 0.64 & 0.65 & 0.69 & 0.67 \\ \midrule
\multicolumn{1}{l|}{\multirow{2}{*}{ZH}} & R$_{\text{pb}}$ & 0.53 & 0.67 & 0.75 & 0.84 & 0.84 \\
\multicolumn{1}{l|}{} & R$_{\text{MT}}$ & 0.50$^{*}$ & 0.54 & 0.64 & 0.74 & 0.75 \\ \bottomrule
\end{tabular}
}
\caption{\sysname performance on reference-based QE.}
\label{tab:nano_qe_w_ref}
\end{table}

\begin{table}[]
\small\centering
  \resizebox{\columnwidth}{!}{
    \begin{tabular}{@{}l|l|lllllll@{}}
        \toprule
         & en-* & en-cs & en-de & en-ja & en-pl & en-ru & en-ta & en-zh \\ \midrule
        COMET & 51.4 & 70.9 & 37.3 & 51.5 & 48.9 & 39.4 & 61.3 & 50.3 \\
        \midrule\midrule
        FT$_\text{\tiny{XL}}$ & 51.4 & 68.7 & 40.6 & 59.6 & 44.3 & 28.2 & 66.3 & 51.8 \\
        \sysname$_\text{\tiny{XL}}$ & 53.8 & 69.5 & 42.7 & 62.6 & 47.1 & 31.5 & 68.4 & 54.8 \\
        FT$_\text{\tiny{XXL}}$ & 57.4 & 71.4 & 47.1 & 65.5 & 52.4 & 36.3 & 70.3 & 58.7   \\
         \sysname$_\text{\tiny{XXL}}$ & 57.6 & 71.8 & 46.6 & 66.3 & 51.0 & 38.5 & 70.4 & 58.8 \\ \bottomrule
    \end{tabular}
    }
    \caption{Segment-level agreement with human ratings for reference-based quality estimation on WMT. }
    \label{tab:wmt_qe_w_ref}
\end{table}

We report \sysname's performance on dialect robustness as the reference-based quality estimation in \autoref{tab:nano_qe_w_ref} and its corresponding WMT performance in \autoref{tab:wmt_qe_w_ref}. In the XL setting, \sysname improves upon both COMET and the finetuning only setup for the dialect robustness and performance on WMT benchmark. However, \sysname achieves comparable performances with finetuning-only setting with XXL models. The findings are consistent with our findings for within-language and reference-free quality estimation settings in the main content: \sysname provides a size-efficient way for models to improve the dialect robustness and their performance on the WMT metrics benchmark.

\section{Dialect Awareness on PT}
\label{app:dialect-awareness-pt}
\begin{table}[t!]
\small\centering
\resizebox{\columnwidth}{!}{
\begin{tabular}{ll|llll}
\toprule
Candidate & Input Tag& FT$_\text{\tiny{XL}}$ & \sysname$_\text{\tiny{XL}}$ & FT$_\text{\tiny{XXL}}$ & \sysname$_\text{\tiny{XXL}}$ \\ \midrule
& perturb & 0.89 & 0.85 & 0.78 & 0.79 \\ \midrule
\multirow{2}{*}{pt-BR} 
& pt-BR &  0.89 & 0.88  & 0.88  & 0.85\\
& pt-PT & 0.88 & 0.87 & 0.88 & 0.93 \\
\midrule
\multirow{2}{*}{pt-PT} 
& pt-BR & 0.85 & 0.84 & 0.85 & 0.84 \\
& pt-PT & 0.84 & 0.84  & 0.85 & 0.91 \\ \bottomrule
\end{tabular}
}
\caption{Dialect Awareness test of \sysname on Portuguese. We score each variant against a translation of English to Portuguese, with the dialect tag as input. }
\vspace{-0.5cm}
\label{tab:awareness_pt}
\end{table}

\autoref{tab:awareness_pt} shows the dialect awareness test of \sysname on Portuguese. As Portuguese and its language variants are not covered in pretraining, we expect \sysname to not perform well in terms of dialect awareness because it has never seen the input dialect tags during training. \autoref{tab:awareness_pt} confirms our expectation. We observe that both finetuning-only and pretraining with \sysname fail to assign higher scores to candidates with matched input language tags over mismatched dialect tags.

\end{document}